\PassOptionsToPackage{table}{xcolor}
\documentclass{filmtogether_report}
\usepackage{graphicx}
\usepackage{booktabs}
\usepackage{multirow}
\usepackage{array}
\usepackage{placeins}
\usepackage{stfloats}
\usepackage{wrapfig}
\usepackage{float}
\usepackage{amsmath,amssymb}

\makeatletter
\newcommand{\fixedsubsection}{\@startsection{subsection}{2}{\z@}%
  {-1.8ex plus -0.5ex minus -.2ex}{0.8ex}{\normalsize\sffamily\bfseries\color{PaperAccent}\raggedright}}
\makeatother

\usepackage{booktabs}
\usepackage{multirow}
\usepackage{graphicx}
\usepackage{xcolor}

\definecolor{tableheader}{HTML}{EDF2F7}
\definecolor{oursrow}{HTML}{EAF5F2}

\usepackage{amsmath,amsfonts,bm}

\def\eqref#1{equation~\ref{#1}}

\def\1{\bm{1}}

\DeclareMathAlphabet{\mathsfit}{\encodingdefault}{\sfdefault}{m}{sl}
\SetMathAlphabet{\mathsfit}{bold}{\encodingdefault}{\sfdefault}{bx}{n}

\usepackage{hyperref}
\usepackage{url}

\title{FilmTogether: Cinematic Video Generation\\from Low-Cost Inputs}
\author{Shiying Zhang\quad Xiaobin Hu\textsuperscript{\dag}\quad Qianchi Zhang\quad Jiaqi Zhao\\[2pt]
Xiaoda Yang\quad Bo Yin\quad Shuicheng Yan}
\affiliations{National University of Singapore\quad
{\footnotesize\textsuperscript{\dag}Project leader}}
\hypersetup{pdfauthor={Shiying Zhang, Xiaobin Hu, Qianchi Zhang, Jiaqi Zhao, Xiaoda Yang, Bo Yin, Shuicheng Yan}}

\begin{document}

\maketitle

\begin{abstract}
The widespread adoption of mobile devices has democratized video recording, but a stark aesthetic divide persists between casual, low-cost captures and high-cost cinematic productions. Bridging this gap requires upgrading heterogeneous production elements, such as lighting, specialized costumes, and materials, without altering the core layout, identity, and temporal motion of the original footage. However, this task is heavily bottlenecked by two important challenges: 1) Lacking a high-quality paired training data specifically designed to handle complex low-to-high appearance shifts. 2) Ignoring the capacity of current video editing models to perceive and enhance the multiple low-cost compositions. To address these issues, we present CineCraft-120K, a large-scale dataset of aligned low- and high-cost video pairs annotated with pair-aware editing prompts. CineCraft-120K features a diverse, high-resolution and high-cost \textbf{Cine}matic \textbf{Craft}ing bank spanning decades of media production, meticulously taxonomized into \textbf{5} broad domains and \textbf{23} fine-grained sub-categories. It also provides element-level mapping between low- and high-cost counterparts, offering pair-aware rich textual prompts that precisely detail the required aesthetic transformations while anchoring structural layout. 
Architecturally, we propose a simple yet effective dual-memory feature adapter, where a text-memory expands semantic capacity for low2high appearance changes, and a hint memory supplies layer-wise visual feature corrections of high-cost element before context injection. Extensive experiments and ablation studies demonstrate the superiority of CineCraft-120K for low-to-high-cost video editing, alongside the effectiveness of our dual-memory adapter.

\par\smallskip
{\small\noindent
Code \& Data:
\href{https://github.com/Shiying-Zhang/FilmTogether}{FilmTogether}
\enspace\textperiodcentered\enspace
\href{https://huggingface.co/datasets/Anna-SY/CineCraft-120K}{CineCraft-120K}\par}
\end{abstract}

\section{Introduction \label{sec:intro}}

The rapid evolution of mobile technology has made video recording accessible to everyone, fundamentally changing how everyday moments are captured and shared. However, a persistent visual gap remains between these casual, low-cost videos and professional cinematic productions. Low-cost videos are typically characterized by amateur framing, ordinary settings, and suboptimal lighting, whereas high-cost productions benefit from professional equipment, sophisticated illumination, and intricate set designs. Advances in denoising, latent, video, and transformer-based diffusion models~\citep{ho2020ddpm,song2020ddim,song2021score,rombach2022latent,ho2022videodiffusion,peebles2023dit} have propelled instruction-based video editing to new heights~\citep{cheng2024consistent,wu2025insvie,bai2025ditto}, yet transforming a basic low-cost video into a cinematic masterpiece remains a largely unexplored frontier. Bridging this aesthetic divide is not a simple global style transfer, but it requires upgrading heterogeneous production elements—such as lighting, specialized costumes, and materials—without altering the core layout, identity, and temporal motion of the original footage.

Realizing this low-to-high cost transformation is heavily bottlenecked by two primary challenges: \textit{1)} Lacking high-quality paired training data for complex appearance shifts. The progress of specialized cinematic editing is hampered by the scarcity of appropriate training data. Current video editing datasets often emphasize generic instructions or broad stylization~\citep{zi2025senorita,openve2025openve,huang2026ffp300k}. They lack the domain-specific mappings and task diversity required to supervise the intricate transition from low-cost to high-cost aesthetics while strictly preserving spatial layout and motion. There is a distinct absence of large-scale, paired data that explicitly captures these element-level production upgrades. \textit{2)} Ignoring the capacity of current models to better perceive and enhance multiple low-cost compositions. At the model level, existing architectures struggle to strengthen the heterogeneous and multiple low-cost compositions.
Standard generators (\textit{e.g.,} Ditto~\citep{bai2025ditto}) typically inject source visual hints directly into a frozen denoising backbone, leaving the model with limited capacity to process semantic instructions related to complex appearance changes, and failing to reliably disentangle the core structure from the low-cost aesthetic.

In this work, we present \textit{CineCraft-120K}, a carefully curated, high-quality dataset comprising over 120K premium cinematic video clips, with over 99\% of the clips feature a resolution of at least 1080\textit{p} and are accompanied by detailed captions. As shown in Fig. 1 and 2, our CineCraft-120K has several characteristics: \textit{1). Superior visual diversity and temporal coverage}: The cinematic crafting bank distinguishes itself through exceptional visual diversity, meticulously taxonomized into 5 broad domains and 23 fine-grained sub-categories, while encompassing over five decades of media production from 1970 to 2023. This extensive temporal coverage prevents the model from overfitting to contemporary trends, ensuring it learns generalized and enduring cinematic aesthetics. \textit{2). High-quality video pairs via a scalable dataset pipeline}: Supported by approximately 1,500 GPU days computational source, it features over 120K high-resolution, aligned low- and high-cost video pairs generated through a highly efficient, automated construction pipeline. Through strict multi-dimensional quality filtering, this intensive approach ensures superior visual richness and clarity, with over 99\% of the resulting cinematic clips featuring 1080p or 4K resolutions. \textit{3). Expressive in pair-aware rich textual captions}: Unlike datasets that provide generic video descriptions, it introduces highly detailed, pair-aware element-wise prompts averaging around 125 dense words. These expressive instructions are grounded in the actual visual transformations between counterparts, precisely detailing coordinated upgrades to heterogeneous elements like lighting, color grading, and materials.

To address the architecture deficit, we propose a dual-memory adapter that seamlessly maps low-cost inputs to cinematic quality with minimal overhead. At the semantic level, the Instruction-Routed Text Memory expands the instruction representation capacity via learnable text-memory token, overcoming the limitations of global textual prompts in detailing diverse production elements. In parallel, the Layer-Wise Hint Memory operates at the visual context injection to supply layer-specific corrections to the source video hints before DiT fusion. This adaptive feature refinement prevents the generative process from being bottlenecked by the coarse textures and defects of the source video, enabling multi-level enhancements from low-level aesthetics to high-level layouts. Synergistically, the text memory deeply enriches the cinematic semantics while the hint memory finely calibrates the visual context, facilitating sophisticated element-level upgrades with high efficiency.

Our contributions are listed as follows: 1). We introduce \textit{CineCraft-120K}, a large-scale, high-quality dataset comprising over 120K aligned low- and high-cost video pairs with pair-aware prompts for cinematic video editing.
2). We propose a Dual-Memory Adapter, which combines text and layer-wise hint memories to seamlessly execute complex aesthetic transformations with minimal overhead.
3). We design a VLM-Aesthetic evaluator and demonstrate the superiority of our framework on the low-to-high-cost video editing task.

\section{CineCraft-120K: Paired Cinematic Data Construction}
\begin{figure*}[t!]
\centering
\includegraphics[width=0.95\textwidth]{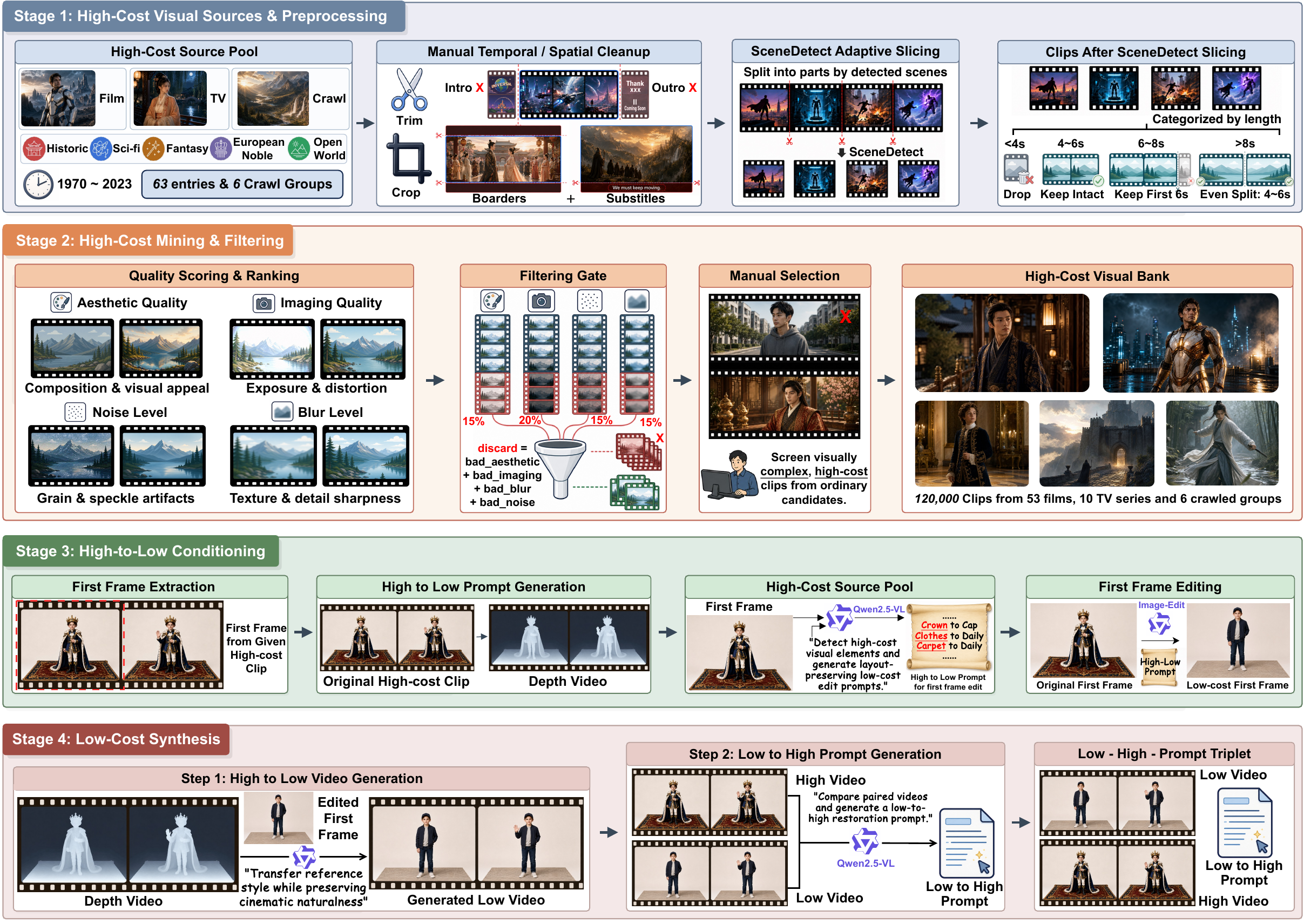}
\vspace{-1mm}
\caption{CineCraft-120K construction pipeline. We curate a high-cost visual bank, synthesize aligned low-cost counterparts through first-frame editing and depth-guided generation, and verify and adapt editing prompts to describe the observed low-to-high transformations.}

\vspace{-1mm}
\label{fig:data-construction-pipeline}
\end{figure*}

We introduce CineCraft-120K, a large-scale paired dataset for upgrading low-cost videos into cinematic productions while preserving their underlying scene structure and motion. Its construction follows three design principles: diverse cinematic scenarios, structure-preserving counterpart synthesis, and pair-aware, element-level instruction caption annotations. Starting from a curated high-cost visual bank, we synthesize low-cost counterparts and subsequently annotate the reverse transformation through precise comparison of each video pair. This approach retains real cinematic footage as the enhancement target while providing explicit supervision for coordinated changes to heterogeneous production elements. Fig. ~\ref{fig:data-construction-pipeline} illustrates the complete pipeline.

\subsection{Cinematic Videos Curation and Pair-Aware Data Construction}
\label{sec:pair-construction}

\noindent\textbf{Diverse Cinematic Source Collection.}
To capture a broad range of cinematic appearances, we collect visually rich footage from films, television series, and crawled open-world videos. The source collection spans media released between 1970 and 2023 and is organized into five broad visual domains and 23 fine-grained categories. This coverage exposes the dataset to varied production designs, lighting styles, costumes, and materials across genres and eras, broadening its scope beyond contemporary cinematic appearances. The crawled subset further extends the collection to visual content beyond narrative film and television.

\noindent\textbf{Shot-Coherent Preprocessing.} We first remove openings, endings, black bars, subtitle regions, and other non-content areas using manually specified temporal boundaries and spatial crops. We then apply adaptive shot detection with PySceneDetect\citep{castellano_pyscenedetect} to obtain temporally coherent clips without crossing detected shot boundaries. Clips shorter than 4 seconds are discarded; those between 4 and 6 seconds are retained; those longer than 6 but shorter than 8 seconds are truncated to their first 6 seconds; and those of at least 8 seconds are evenly divided into approximately 4–6-second segments. This procedure produces compact source clips while avoiding short residual segments and preserving within-shot continuity.

\noindent\textbf{Multi-Dimensional Quality Curation.} 
We combine manual screening for visually rich, high-production-value content with automated perceptual quality filtering. VBench \citep{huang2024vbench} evaluates aesthetic and imaging quality, for which we reject the bottom 15\% and 20\% of candidates, respectively. FineVQ \citep{duan2025finevq} provides complementary blur- and noise-related quality assessments, with the lowest-quality 15\% rejected under each criterion. A clip is removed if it falls into the rejection set of any criterion. This filtering strategy screens for complementary failure modes, including weak composition, exposure and distortion artifacts, excessive noise, and blurred details. The retained clips form the high-cost visual bank that supplies cinematic targets for subsequent pair construction.

\noindent\textbf{Structure-Guided High-to-Low Synthesis.}
To construct paired supervision, we generate a low-cost counterpart for each selected high-cost clip by simplifying its production-related appearance under structural guidance. 
From each reference clip, we extract the first frame and estimate a depth sequence using Video-Depth-Anything \citep{chen2025videodepthanything}. Qwen2.5-VL-32B-Instruct \citep{bai2025qwen25vl} examines the first frame and generates a high-to-low editing instruction that identifies visually elaborate elements and specifies replacements while retaining their spatial arrangement. Qwen-Image-Edit \citep{wu2025qwenimage}, implemented through DiffSynth-Studio, applies this instruction to produce a low-cost first frame. Wan2.1-VACE-14B \citep{jiang2025vace,wan2025} then synthesizes the low-cost video conditioned on the edited frame and reference depth sequence. The edited frame establishes the intended low-cost appearance, while the depth sequence guides scene geometry and temporal motion.
The overall construction effort uses approximately 1,500 GPU-days. This process supports large-scale pair generation with appearance changes guided by a shared underlying scene structure. 

\begin{figure*}[t!]
\centering
\includegraphics[width=0.95\textwidth]{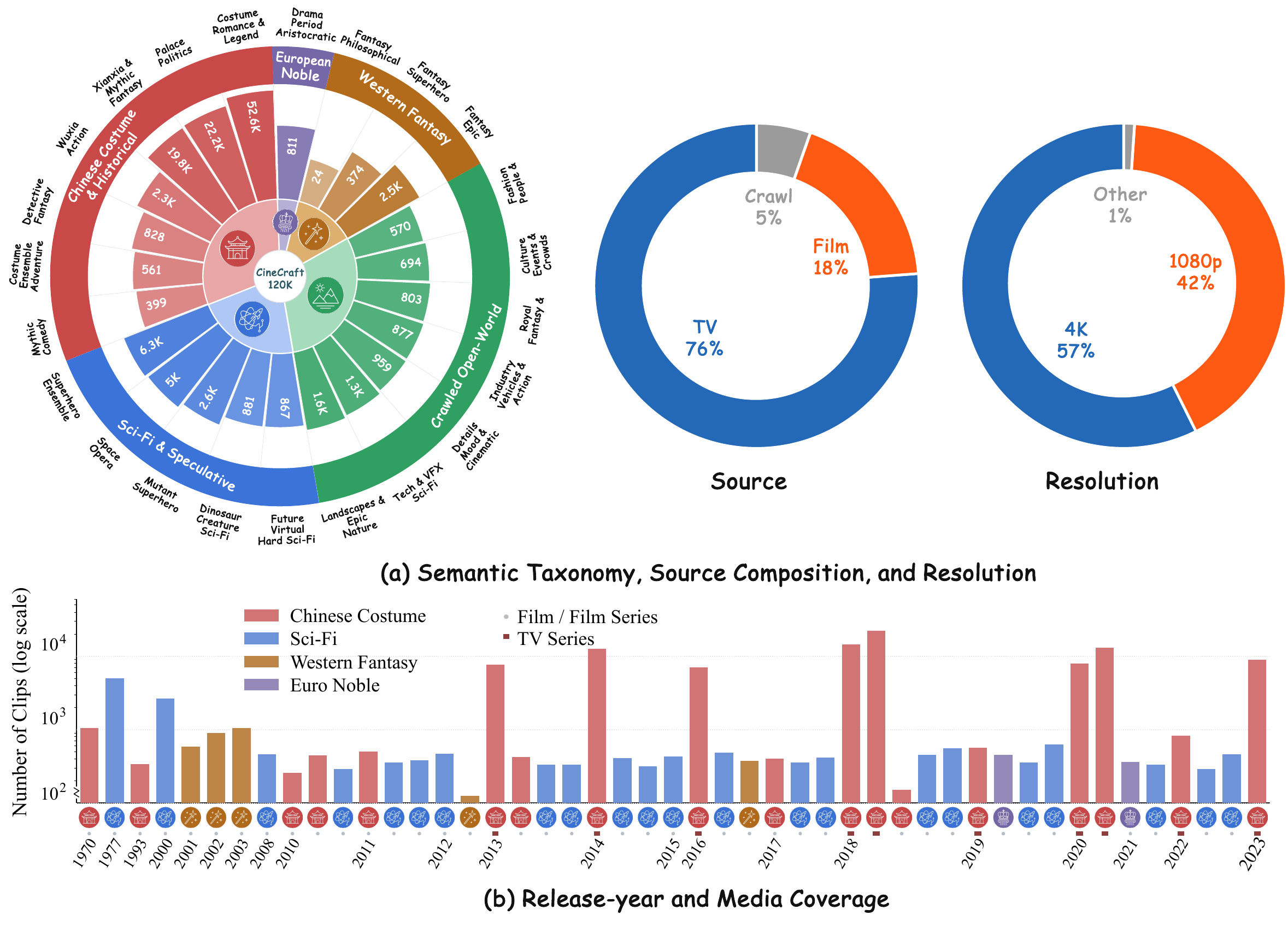}
\vspace{-1mm}
\caption{Source composition and visual coverage of CineCraft-120K.}
\label{fig:source-visual-coverage}
\vspace{-1mm}
\end{figure*}

\noindent\textbf{Prompt-Guided Verification and Adaptive Annotation.}
We use the preceding high-to-low editing instruction as a reference to verify whether the synthesized low-cost video realizes the intended appearance changes. Specifically, {Qwen3-VL-8B-Instruct} jointly examines the editing instruction, the original high-cost video, and its generated low-cost counterpart to assess whether the specified element-level modifications are reflected in the resulting pair. This comparison identifies discrepancies between the intended and realized edits, including changes that are missing, only partially achieved, or introduced beyond the instruction.

When discrepancies are detected, the instruction is adaptively revised to reflect the transformations actually observed in the video pair. The verified and adjusted transformation description is then expressed as a low-to-high enhancement instruction, specifying how to upgrade the generated low-cost video toward its high-cost reference. This verification-and-adaptation process grounds the final supervision in both the original editing intent and the realized visual changes, reducing mismatches between textual instructions and paired video content.

\begin{figure*}[t!]
\centering
\includegraphics[width=0.95\textwidth]{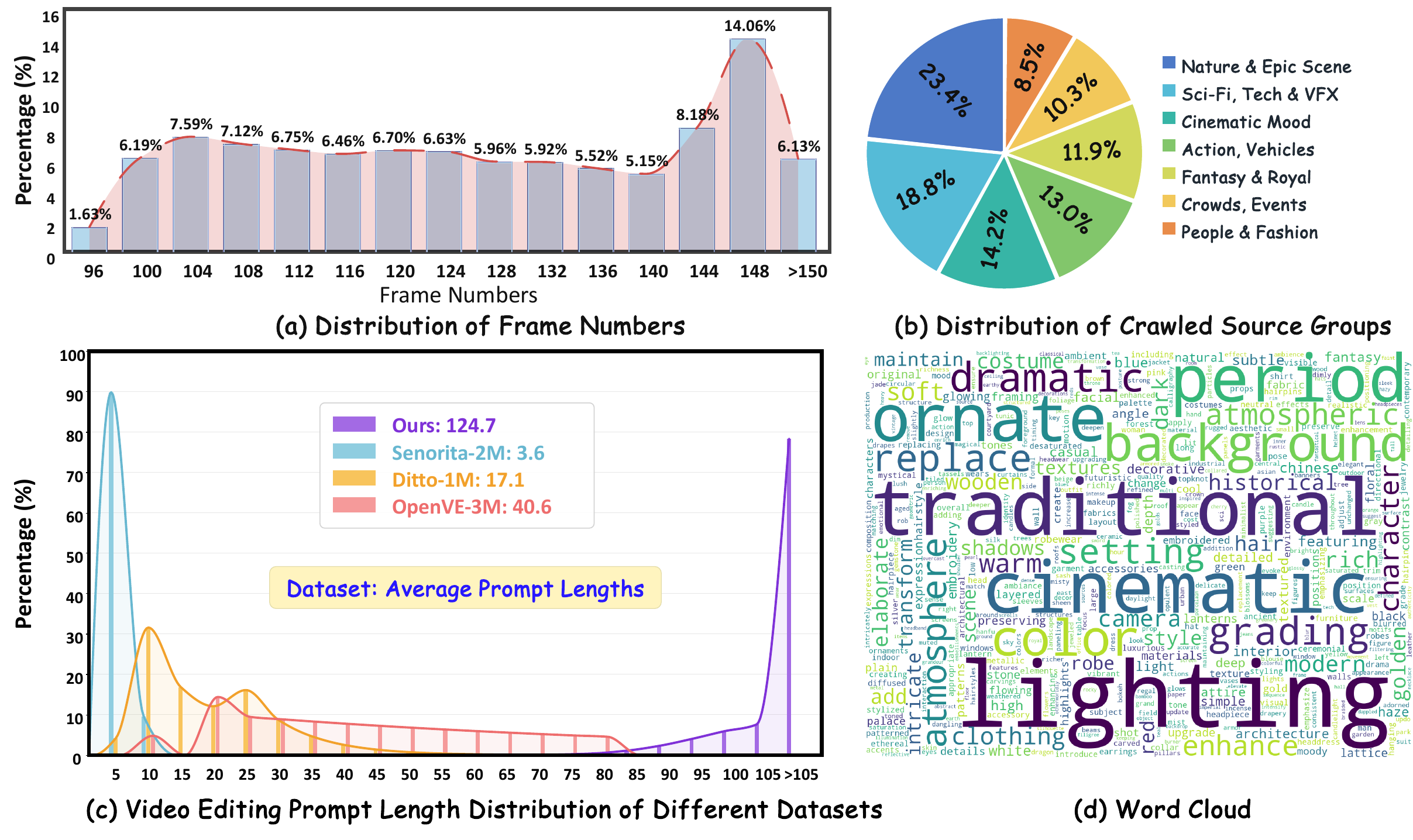}
\vspace{-1mm}
\caption{CineCraft-120K statistics: (a) source-clip frame-count distribution; (b) crawled-content categories; (c) editing-prompt length distributions; and (d) editing-instruction word cloud.}
\label{fig:crawl-video-text-statistics}
\vspace{-1mm}
\end{figure*}

\subsection{Dataset Characteristics: Diversity, Quality, and Expressiveness}
\label{sec:dataset-statistics}

We characterize CineCraft-120K through visual coverage, source quality, and
pair-aware instructions. Its high-cost bank contains \textbf{124,902 clips}
totaling \textbf{176.38 hours}; Figures~\ref{fig:source-visual-coverage} and
\ref{fig:crawl-video-text-statistics} summarize its statistics.

\noindent\textbf{Visual Diversity and Historical Coverage.}
CineCraft-120K spans 5 broad domains and 23 fine-grained categories. The
taxonomy covers Chinese costume and historical scenes (7 categories), sci-fi
and speculative content (5), Western fantasy (3), European aristocratic scenes,
and crawled open-world content (7), from wuxia and palace settings to space
opera and epic landscapes (Fig.~\ref{fig:source-visual-coverage}(a)).

Sources span 1970--2023, while crawled clips extend beyond film and television.
Their largest visual groups are nature/epic scenes (23.4\%), sci-fi/technology/VFX
(18.8\%), cinematic mood (14.2\%), and action/vehicles (13.0\%); the remaining
groups are fantasy/royalty (11.9\%), crowds/events (10.3\%), and people/fashion
(8.5\%) (Figs.~\ref{fig:source-visual-coverage}(b) and
\ref{fig:crawl-video-text-statistics}(b)).

\noindent\textbf{Large-Scale, High-Quality Cinematic Paired Videos.}
Television contributes about 76\% of the visual bank, and 99\% of source clips
are 1080p or 4K. Combined with multi-dimensional quality filtering, these
sources provide detailed appearance references; Fig.~\ref{fig:crawl-video-text-statistics}(a)
shows their frame-count distribution.

\noindent\textbf{Expressive, Pair-Aware Editing Captions.}
Prompt-guided verification grounds each low-to-high instruction in the edits
realized by its video pair. Instructions average \textbf{124.7 words} (median
\textbf{122}; IQR \textbf{108--138}) and describe coordinated changes to
lighting, color, clothing, materials, and scene appearance. Their length and
vocabulary distributions appear in Figs.~\ref{fig:crawl-video-text-statistics}(c,d).

\section{Methodology: Dual-Memory Adaptation}
Given a low-cost source video and an editing instruction, FilmTogether aims to enhance cinematic appearance while preserving source identity, spatial layout, and temporal motion. This task requires coordinated changes to heterogeneous production elements, including illumination, costumes, materials, and set design. It therefore places complementary demands on the conditioning pathways: textual features must represent the intended appearance changes, while visual features must retain structural guidance without restricting generation to the source appearance.

FilmTogether builds on Wan2.1-VACE-14B \citep{wan2025,jiang2025vace}.

\begin{figure}[t!]
    \centering
    \includegraphics[width=0.95\textwidth,trim=108 52 108 4,clip]{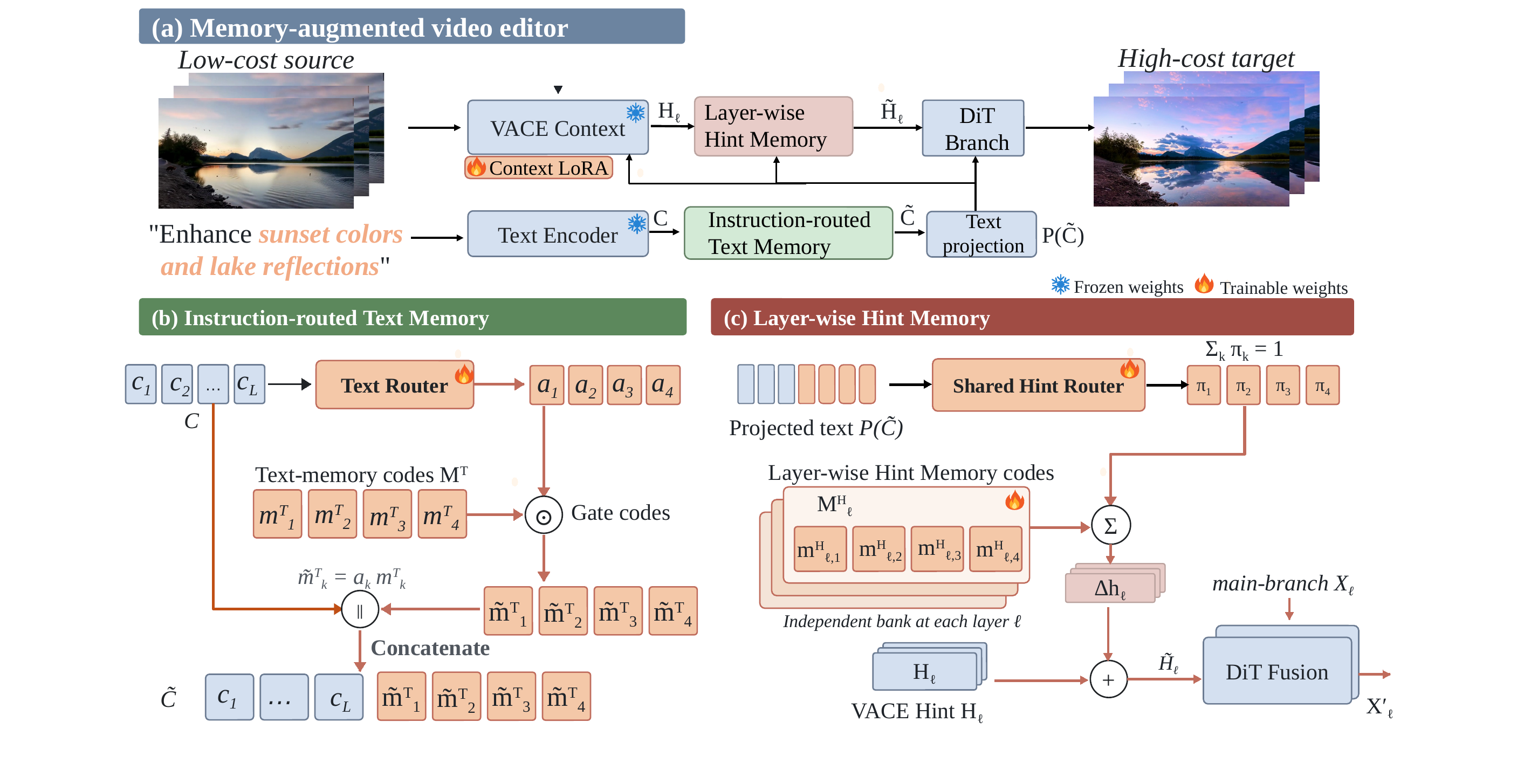}
    \vspace{-1mm}
    \caption{FilmTogether architecture: (a) overall framework; (b) instruction-routed Text Memory enriches semantic conditioning; and (c) layer-wise Hint Memory refines source visual hints before DiT fusion.}
\label{fig:model-training-pipeline}
\end{figure}

As shown in Fig.~\ref{fig:model-training-pipeline}, we introduce a \textbf{dual-memory adapter} that augments both conditioning pathways while keeping the pretrained backbone frozen.

\emph{Instruction-Routed Text Memory} supplements the instruction with learnable, adaptively gated tokens.

\emph{Layer-Wise Hint Memory} uses the augmented instruction to retrieve layer-specific residuals that refine source-video hints before their injection into the transformer-based DiT Main Branch~\citep{vaswani2017attention,peebles2023dit}.

\noindent\textbf{Instruction-Routed Text Memory.}
The frozen text encoder provides general-purpose instruction features but cannot directly adapt its parameters to cinematic enhancement. We introduce a learnable memory that expands the conditioning representation while retaining the original instruction tokens.

For clarity, let $\mathbf{C}=[\mathbf{c}_1,\ldots,\mathbf{c}_L]\in
\mathbb{R}^{B\times L\times d_T}$ denote the frozen text-encoder features and
$\bar{\mathbf{c}}=L^{-1}\sum_i\mathbf{c}_i$ their pooled representation.  
The learnable text-memory codes are
$\mathbf{M}^T=\{\mathbf{m}^{T}_{k}\}_{k=1}^{K}$ ($K=4$), each with the same dimension as a text token. 
A lightweight router predicts independent sigmoid gates:
\begin{equation}
    \mathbf{a}^{T}=\operatorname{sigmoid}\!\left(
    \mathbf{W}^{T}_{2}\,\phi\!\left(
    \mathbf{W}^{T}_{1}\operatorname{LN}(\bar{\mathbf{c}})
    \right)\right),
    \qquad 0\leq a_k^T\leq1.
\end{equation}
\textcolor{black}{where $\bar{\mathbf c}$ is the mean-pooled instruction feature, $\operatorname{LN}$ is layer normalization~\citep{ba2016layer}, $\phi$ is GELU~\citep{hendrycks2016gelu}, $\operatorname{sigmoid}$ independently maps each router logit to $[0,1]$, $\mathbf W_1^T$ and $\mathbf W_2^T$ are the trainable Text-Router projections, $k$ indexes a memory code, and the superscript $T$ denotes Text Memory.}
The gates scale the codes before appending them to the original text sequence:
\begin{equation}
    \widetilde{\mathbf{C}}=
    [\mathbf{C};a^{T}_{1}\mathbf{m}^{T}_{1};\ldots;
    a^{T}_{K}\mathbf{m}^{T}_{K}].
\end{equation}
\textcolor{black}{where $\widetilde{\mathbf C}\in\mathbb R^{B\times(L+K)\times d_T}$ is the augmented instruction sequence, $B$ is the batch size, $L$ is the instruction length, $d_T$ is the text-feature dimension, $\mathbf c_i\in\mathbb R^{d_T}$ is the $i$-th frozen text token, $K=4$ is the number of learnable codes, $\mathbf m_k^T\in\mathbb R^{d_T}$ is the $k$-th code, $[\,;\,]$ denotes token-wise concatenation. 
}

Independently gated memory tokens expand semantic capacity for coordinated appearance upgrades while retaining the original instruction tokens.
\textcolor{black}{Unlike softmax selection, the independent gates can activate several codes simultaneously for instructions that combine lighting, material, costume, and set-design changes, while appending the codes preserves every literal instruction token.}

\noindent\textbf{Layer-Wise Hint Memory.}
The VACE Context Branch extracts source-video hints
$\{H_\ell\}$ at designated fusion locations. These features carry structural information useful for preserving the source, but may also encode its low-cost textures, lighting, and other appearance characteristics. Hint Memory introduces an instruction-conditioned residual at each fusion location, allowing the visual conditioning to adapt before entering the main branch.
\begin{equation}
    \boldsymbol{\pi}^{H}=\operatorname{softmax}\!\left(
    \mathbf{W}^{H}_{2}\,\phi\!\left(
    \mathbf{W}^{H}_{1}\operatorname{LN}
    (\operatorname{Pool}(P(\widetilde{\mathbf{C}})))
    \right)\right),
    \qquad \sum_{k=1}^{K}\pi_k^H=1.
\end{equation}
\textcolor{black}{where $P$ is the frozen projection from text width $d_T$ to the DiT hidden width, $\operatorname{Pool}$ averages the projected tokens, $\mathbf W_1^H$ and $\mathbf W_2^H$ are the trainable Hint-Router projections, 
$\boldsymbol\pi^H=[\pi_1^H,\ldots,\pi_K^H]$ is the resulting instruction-conditioned distribution, and the superscript $H$ denotes Hint Memory.}
Each fusion location $\ell$ owns an independent bank
$\mathbf{M}_{\ell}^{H}=\{\mathbf{m}^{H}_{\ell,k}\}_{k=1}^{K}$ in the DiT
hidden dimension. The shared routing decision reads a layer-specific correction:
\begin{equation}
    \Delta\mathbf{h}_{\ell}=
    \sum_{k=1}^{K}\pi_k^H\mathbf{m}^{H}_{\ell,k}.
\end{equation}
\textcolor{black}{where $\ell$ indexes 
a DiT
fusion location, 
$\mathbf m_{\ell,k}^H$ is the $k$-th code in that bank, $\pi_k^H$ is its shared instruction-dependent weight, and $\Delta\mathbf h_\ell$ is the resulting layer-specific correction.}

Here the correction is broadcast to all hint tokens. Independent token banks let early, middle, and late fusion locations learn different corrections. 
Its interaction with position-dependent features and subsequent attention yields
non-uniform spatiotemporal effects. 
Placing the correction before fusion is important, after considering the fact that 
the original hint continues to anchor identity, geometry, and motion, while the routed residual removes the need for the main branch to inherit its low-cost appearance unchanged. Layer-specific
token banks further allow early injections to refine local texture and illumination and later injections to
adjust higher-level scene organization, without unfreezing the visual backbone.

\noindent\textbf{Distilled Aesthetic Evaluator.}
Following knowledge distillation~\citep{hinton2015distilling}, we distill Gemini 3.1 Pro Preview's aesthetic assessments of 5,000 edited clips into Qwen3-VL-8B-Thinking~\citep{bai2025qwen3vl} using LoRA~\citep{hu2022lora} and token-level cross-entropy loss. Supervision includes five aspect scores, an overall score, and corresponding rationales. Used only for evaluation, the frozen model receives edited frames and weak instruction context. We report its overall score as VLM-Aesthetic; Appendix~\ref{sec:aesthetic-evaluator-details} provides the full rubric.

\section{Experiments and Results}
\newcommand{\bestscore}[1]{\textbf{#1}}

\subsection{Experimental Setup}
\label{sec:experiment-setup}

\noindent\textbf{Implementation details.}
Our model builds on Wan2.1-VACE-14B~\citep{wan2025,jiang2025vace} with Ditto's Context-Branch LoRA~\citep{bai2025ditto}. We augment the pretrained editor with Text Memory and layer-wise Hint Memory, then fine-tune on the CineCraft-120K training split. The reference-frame prior is linearly annealed and disabled at inference.
We use
AdamW~\citep{loshchilov2019decoupled} with a learning rate of $5\times10^{-5}$
and weight decay of $0.01$. Both stages use bfloat16 and an effective batch size of eight. 

\noindent\textbf{Competitors.}
We compare \textit{Filmtogether} against VACE~\citep{jiang2025vace}, Bernini-R-14B
\citep{team2026bernini}, Ditto~\citep{bai2025ditto}, InsV2V~\citep{cheng2024consistent},
InsViE~\citep{wu2025insvie}, TokenFlow~\citep{geyer2024tokenflow},
ICVE~\citep{liao2025context}, Kiwi-Edit~\citep{kiwiedit},
SANA-Streaming~\citep{zhao2026sana}, Se{\~n}orita~\citep{zi2025senorita},
ReCo~\citep{zhang2025reco}, and Lucy-Edit~\citep{decart2025lucyedit}.  For a fair comparison, we evaluate each baseline using the settings recommended in its paper or official implementation.

\noindent\textbf{Evaluation metrics.}
We measure temporal consistency using CLIP-F~\citep{radford2021learning} and DINO-F~\citep{caron2021emerging}, which compute mean adjacent-frame feature cosine similarity, and text-conditioned preference using frame-averaged PickScore~\citep{kirstain2023pick}. For cinematic quality, we use the VLM-Aesthetic evaluator detailed in Appendix~\ref{sec:aesthetic-evaluator-details} to deterministically score two edited frames sampled at 1 fps and resized to 640 pixels wide. We report its holistic score, a semantic-design score averaging narrative/emotional fit, style/world consistency, and composition/lighting, and separate color/texture and hierarchy/readability scores.

\noindent\textbf{Cinematic Crafting Benchmark (CineCraft-Bench).}
CineCraft-Bench comprises 100 videos spanning Chinese costume and historical scenes, science fiction, Western fantasy, European aristocratic settings, and open-world content. These videos cover diverse subjects, environments, and cinematic styles, with editing instructions targeting lighting, color grading, costumes, materials, and scene appearance.

\subsection{Quantitative Comparison}
\label{sec:quantitative_comparison}

Table~\ref{tab:sota-comparison} compares FilmTogether with 12 state-of-the-art video editing models on low-to-high-cost cinematic enhancement. FilmTogether achieves the highest CLIP-F (0.9967) and DINO-F (0.9985), indicating strong temporal consistency during appearance transformation. Its VLM-Aesthetic score is 3.87, compared with 3.77 for Bernini-R-14B.

The fine-grained aesthetic scores further highlight its effectiveness for low-to-high enhancement. FilmTogether leads in narrative/style/lighting (3.88), color/texture (3.86), and hierarchy/readability (3.91), respectively. These results support its ability to coordinate improvements across multiple production elements while maintaining temporally consistent outputs. Overall, the results demonstrate the method's strength in cinematic appearance enhancement and temporal consistency, the two complementary requirements of low-to-high-cost video editing.

\begin{figure*}[t!]
\centering
\includegraphics[width=0.95\textwidth]{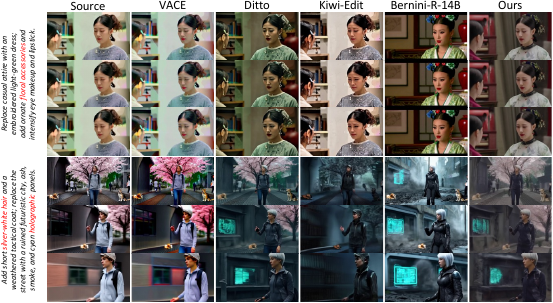}
\vspace{-1mm}
\caption{Qualitative comparison on two representative CineCraft-Bench cases.
Each method receives the same source video and editing instruction, and three
matched timestamps are shown. FilmTogether realizes the requested period
costume and science-fiction transformations while preserving subject identity,
pose, and temporal structure.}
\label{fig:qualitative-method-comparison}
\vspace{-1mm}
\end{figure*}

\fixedsubsection{Qualitative Comparison}
\label{sec:qualitative-comparison}

We inspect representative results under matched inputs and instructions.
Figure~\ref{fig:qualitative-method-comparison} presents two representative
cases and exposes the trade-off between edit strength and source preservation.
VACE often under-edits, while stronger baselines can alter faces or scene
layout. FilmTogether introduces period-specific portrait cues and futuristic
holographic panels and smoke while retaining identity, pose, and viewpoint
across frames, yielding controlled enhancement rather than unconstrained
redrawing.
Additional examples of cinematic editing on everyday basketball and piano videos are provided in Appendix~\ref{sec:everyday-video-editing}.

\begin{table}[!htbp]
\centering
\small
\caption{Quantitative comparison with 12 state-of-the-art video editing models. Best and second-best results are bold and underlined, respectively. All metrics: higher is better.
Narr./Style/Lighting averages narrative/emotional fit,
style/world consistency, and composition/lighting.
}
\label{tab:sota-comparison}
\setlength{\tabcolsep}{3pt}
\renewcommand{\arraystretch}{1.05}
\resizebox{\textwidth}{!}{%
\begin{tabular}{@{}lccccccc@{}}
\toprule
\rowcolor{tableheader}
\textbf{Method} &
\textbf{CLIP-F} &
\textbf{DINO-F} &
\textbf{PickScore} &
\shortstack{\textbf{VLM}\\\textbf{Aesthetic}} &
\shortstack{\textbf{Narr./Style/}\\\textbf{Lighting}} &
\shortstack{\textbf{Color/}\\\textbf{Texture}} &
\shortstack{\textbf{Hierarchy/}\\\textbf{Readability}} \\
\midrule

InsV2V (2024)
& 0.9857 & 0.9865 & 18.02
& 1.89 & 1.95 & 1.84 & 2.16 \\

TokenFlow (2024)
& 0.9935 & 0.9956 & 18.85
& 2.98 & 2.99 & 2.96 & 3.18 \\

VACE (2025)
& 0.9884 & 0.9896 & 18.09
& 2.31 & 2.34 & 2.26 & 2.53 \\

InsViE (2025)
& 0.9918 & 0.9923 & 18.25
& 1.89 & 2.09 & 1.71 & 2.03 \\

ICVE (2025)
& 0.9933 & 0.9976 & 17.94
& 2.53 & 2.55 & 2.52 & 2.84 \\

Se{\~n}orita (2025)
& 0.9953 & \underline{0.9980} & 18.18
& 2.54 & 2.55 & 2.52 & 2.68 \\

ReCo (2025)
& \underline{0.9962} & 0.9977 & \underline{18.88}
& 3.44 & 3.47 & 3.40 & 3.43 \\

Lucy-Edit (2025)
& 0.9946 & 0.9977 & 18.65
& 3.59 & 3.59 & 3.58 & 3.63 \\

Bernini-R-14B (2026)
& 0.9903 & 0.9896 & \textbf{19.05}
& \underline{3.77} & \underline{3.80}
& \underline{3.74} & \underline{3.80} \\

Ditto (2026)
& 0.9911 & 0.9913 & 18.83
& 3.22 & 3.26 & 3.18 & 3.42 \\

Kiwi-Edit (2026)
& 0.9931 & 0.9958 & 17.79
& 2.28 & 2.28 & 2.28 & 2.31 \\

SANA-Streaming (2026)
& 0.9937 & 0.9964 & 18.06
& 2.33 & 2.40 & 2.28 & 2.50 \\

\midrule
\rowcolor{oursrow}
\textbf{FilmTogether (Ours)}
& \textbf{0.9967} & \textbf{0.9985} & 18.83
& \textbf{3.87} & \textbf{3.88}
& \textbf{3.86} & \textbf{3.91} \\

\bottomrule
\end{tabular}%
}
\vspace{-3mm}
\end{table}

\subsection{Dataset Ablation}
\label{sec:dataset-ablation}
The paired CineCraft-120K data improves VLM-Aesthetic from 3.22 to 3.78 for
Ditto and from 3.66 to 3.87 for FilmTogether, while also improving temporal
and fine-grained appearance scores (Table~\ref{tab:model-data-ablation}).

\begin{table}[!htbp]
\centering
\caption{Effectiveness of CineCraft-120K across two backbones. Best results
within each backbone are bold. Green rows use CineCraft-120K. All metrics are
higher-is-better.}
\label{tab:model-data-ablation}
\setlength{\tabcolsep}{3pt}
\renewcommand{\arraystretch}{1.05}
\resizebox{\textwidth}{!}{%
\begin{tabular}{@{}llccccccc@{}}
\toprule
\rowcolor{tableheader}
\textbf{Backbone} & \textbf{Dataset} & \textbf{CLIP-F} & \textbf{DINO-F} &
\textbf{PickScore} & \shortstack{\textbf{VLM}\\\textbf{Aesthetic}} &
\shortstack{\textbf{Narr./Style/}\\\textbf{Lighting}} &
\shortstack{\textbf{Color/}\\\textbf{Texture}} &
\shortstack{\textbf{Hierarchy/}\\\textbf{Readability}} \\
\midrule
\multirow{2}{*}{Ditto} & Ditto Dataset & 0.9911 & 0.9913 & 18.83 & 3.22 & 3.26 & 3.18 & 3.42 \\
& \cellcolor{oursrow}CineCraft-120K & \cellcolor{oursrow}\textbf{0.9968} &
\cellcolor{oursrow}\textbf{0.9985} & \cellcolor{oursrow}\textbf{18.87} &
\cellcolor{oursrow}\textbf{3.78} & \cellcolor{oursrow}\textbf{3.80} &
\cellcolor{oursrow}\textbf{3.76} & \cellcolor{oursrow}\textbf{3.82} \\
\midrule
\multirow{2}{*}{FilmTogether} & Ditto Dataset & 0.9917 & 0.9920 & \textbf{19.04} &
3.66 & 3.71 & 3.62 & 3.75 \\
& \cellcolor{oursrow}CineCraft-120K & \cellcolor{oursrow}\textbf{0.9967} &
\cellcolor{oursrow}\textbf{0.9985} & \cellcolor{oursrow}18.83 &
\cellcolor{oursrow}\textbf{3.87} & \cellcolor{oursrow}\textbf{3.88} &
\cellcolor{oursrow}\textbf{3.86} & \cellcolor{oursrow}\textbf{3.91} \\
\bottomrule
\end{tabular}%
}
\end{table}

\subsection{User Study}
\label{sec:user-study}

\noindent\textbf{User Study Protocol.}
Ten participants compare \textit{FilmTogether} with Ditto and Bernini-R-14B on ten randomly selected low-to-high editing cases. For each case, they view the source video, editing instruction, and three synchronized outputs with anonymized method labels. Case order is randomized across participants, and output positions vary across cases. Participants select one preferred output for each of six criteria: instruction following, cinematic enhancement, content preservation, motion fidelity, temporal consistency, and visual quality and realism, yielding 600 selections in total and 100 per criterion. We report selection rates for each criterion and their unweighted mean.

\begin{figure}[t!]
\centering
\includegraphics[width=0.95\linewidth]{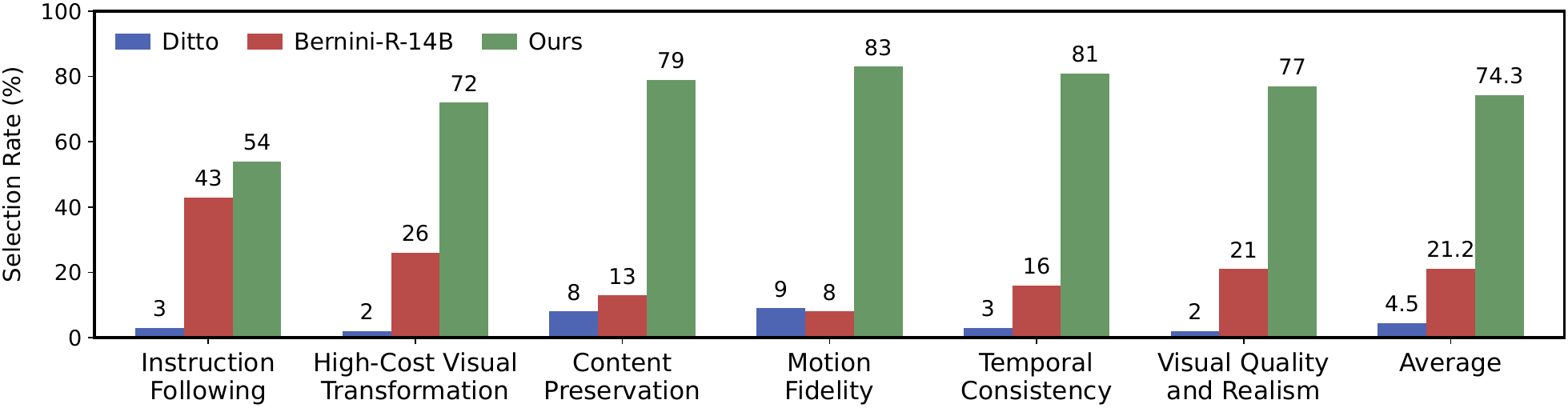}
\vspace{-2mm}
\caption{Human preference for low-to-high video editing. Values are three-way
selection rates (\%, $\uparrow$), with 100 responses per criterion from ten
participants evaluating ten cases. Average is the unweighted mean over six
criteria.}
\label{fig:user-study}
\vspace{-1mm}
\end{figure}

\noindent\textbf{Human Preference Results.}
Figure~\ref{fig:user-study} shows that \textit{FilmTogether} achieves the highest observed selection rate across all six criteria, averaging 74.3\%, compared with 21.2\% for Bernini-R-14B and 4.5\% for Ditto. It receives 72\% of selections for cinematic enhancement, compared with 26\% for Bernini-R-14B and 2\% for Ditto. It also leads in content preservation (79\%), motion fidelity (83\%), temporal consistency (81\%), and visual quality and realism (77\%). The margin is smaller for instruction following (54\% versus 43\% for Bernini-R-14B). These results support FilmTogether's perceived ability to enhance cinematic appearance while retaining source content and motion on the evaluated cases.

\section{Conclusion}
\label{sec:conclusion}

We present FilmTogether for transforming low-cost videos into cinematic outputs
while preserving source content and motion. CineCraft-120K provides 23 diverse
scenarios, aligned video pairs, and pair-aware editing instructions. Its
dual-memory adapter coordinates semantic edits with layer-wise visual guidance.
Quantitative results and human evaluation show improved cinematic appearance
and temporal consistency, supporting accessible low-to-high video enhancement.

\FloatBarrier

\subsection*{AI Use Statement}
Generative AI tools assisted with manuscript language editing. The authors
reviewed all assisted edits and take responsibility for the final content;
scientific contributions, methods, results, and conclusions are the authors' own.

\appendix
\renewcommand{\thefigure}{A\arabic{figure}}
\renewcommand{\theHfigure}{appendix.A\arabic{figure}}
\setcounter{figure}{0}
\renewcommand{\thetable}{A\arabic{table}}
\renewcommand{\theHtable}{appendix.A\arabic{table}}
\setcounter{table}{0}

\section{Related Work}
\label{sec:related-work}

\noindent\textbf{Video-to-video Datasets.}
Large-scale captioned video corpora and caption-transfer pipelines, including WebVid, VideoCC, Panda-70M, InternVid, and OpenVid-1M, support video-language pretraining but do not provide paired editing transformations~\citep{bain2021frozen,nagrani2022learning,chen2024panda,wang2024internvid,nan2025openvid}. InsV2V synthesizes video pairs from related prompts~\citep{cheng2024consistent}, whereas InsViE-1M, Ditto-1M, and FFP-300K use frame-guided editing and temporal propagation or synthesis~\citep{wu2025insvie,bai2025ditto,huang2026ffp300k}. Se{\~n}orita-2M and OpenVE-3M employ specialist or category-specific pipelines for diverse editing operations~\citep{zi2025senorita,openve2025openve}.
Appendix Table~\ref{tab:related-video-datasets} summarizes the key characteristics of these approaches.
Our CineCraft-120K specializes this strategy for cinematic enhancement, pairing curated high-production-value targets with synthesized low-cost counterparts and prompts grounded in observed differences. It focuses on coordinated upgrades to lighting, costumes, and materials while preserving scene structure and motion.

\begin{table}[h!]
\centering
\caption{Comparison of video-editing datasets and the CineCraft-120K high-cost visual bank.}
\label{tab:related-video-datasets}
\fontsize{8}{9.6}\selectfont
\setlength{\tabcolsep}{2pt}
\renewcommand{\arraystretch}{1.0}
\newcommand{\rwDataset}[2]{#1~\citep{#2}}
\newcommand{\rwTableContent}{%
\toprule
\textbf{Dataset} & \textbf{Scale} & \textbf{Editing objective} & \textbf{Target provenance} & \textbf{Pair construction} \\
\midrule
\rwDataset{InsV2V}{cheng2024consistent} & 68.4k & Object/scene/style edits & Synthetic & Video Prompt-to-Prompt \\
\rwDataset{InsViE-1M}{wu2025insvie} & 1.02M & Local/global edits & Generated/image-derived & Frame propagation + images \\
\rwDataset{Se\~norita-2M}{zi2025senorita} & 2M & Edits + translation & Edited/original/derived & Specialists + transforms \\
\rwDataset{Ditto-1M}{bai2025ditto} & 1M & Local/global edits & Generated & Frame edit + depth synthesis \\
\rwDataset{OpenVE-3M}{openve2025openve} & 3M & Multi-task editing & Generated/original & Category-specific \\
\rwDataset{FFP-300K}{huang2026ffp300k} & 300k & Inpainting/stylization & Generated & Inpainting + stylization \\
\midrule
CineCraft-120K (Ours) & 124.9k & Cinematic enhancement & Curated high-cost videos & High-to-low synthesis \\
\bottomrule
}
\resizebox{\textwidth}{!}{%
\begin{tabular}{@{}lrlll@{}}
\rwTableContent
\end{tabular}%
}
\end{table}

\noindent\textbf{Instruction-based Video Editing.}
Diffusion-based video editors adapt image priors through one-shot tuning, zero-shot inversion and attention reuse, anchor-frame propagation, or flow-guided attention~\citep{wu2023tuneavideo,khachatryan2023text2videozero,qi2023fatezero,ceylan2023pix2video,liu2024videop2p,cong2024flatten}. AnyV2V instead combines an image editor with image-to-video propagation in a tuning-free pipeline~\citep{ku2024anyv2v}. Inversion-based methods such as TokenFlow~\citep{geyer2024tokenflow} reuse source features and inter-frame correspondences under target text guidance, while paired-data methods such as InsV2V~\citep{cheng2024consistent} and InsViE~\citep{wu2025insvie} learn instruction-driven transformations. FFP-300K~\citep{huang2026ffp300k} uses an edited first frame to specify the target appearance. Ditto's Editto~\citep{bai2025ditto} bridges reference-guided training and instruction-driven inference by gradually removing this reference during training. OpenVE-Edit~\citep{openve2025openve} further enriches conditioning through a video--instruction encoder and a task-aware MOE connector. Our dual-memory adapter augments text conditioning and refines layer-wise source hints within the existing editor, supporting cinematic instruction interpretation and low-cost appearance adaptation.

\FloatBarrier

\section{Additional Experimental Details}
\subsection{Aesthetic Evaluator Details}
\label{sec:aesthetic-evaluator-details}

\paragraph{Scoring rubric.}
The teacher and distilled evaluator use a four-level rubric for every aspect:
4 denotes an excellent or strongly successful result, 3 a good result with
minor issues, 2 a weak result with clear issues, and 1 a failed result or one
that harms the intended aesthetic. The five aspects are defined as follows.
\begin{enumerate}
    \item \textbf{Narrative and emotional fit}: integration with scene mood,
    emotional tone, and visual context without distracting anomalies.
    \item \textbf{Style and world consistency}: compatibility with the era,
    genre, environment, and overall visual language.
    \item \textbf{Composition and lighting design}: composition, contrast,
    lighting hierarchy, lens design, and shot-level arrangement.
    \item \textbf{Color and texture refinement}: color harmony, saturation
    control, material quality, and avoidance of cheap or generic filter looks.
    \item \textbf{Visual hierarchy and readability}: clarity of the focal
    hierarchy and visibility of important content.
\end{enumerate}
The holistic score is produced independently in $[1,4]$ and is not defined as
the arithmetic mean of the five aspect scores. Utility edits such as removal,
cleanup, and denoising are not penalized for being visually subtle when the
result is seamless and natural.

\paragraph{Student input and output.}
At evaluation time, frames are extracted from the edited video at 1 frame per
second, resized to 640 pixels in width, and the first two available frames are
used. The source video and high-cost target are excluded.
The instruction is supplied only as weak context for the intended visual
direction. The evaluator must emit a valid JSON object containing the five
integer aspect scores, a floating-point holistic score, an uncertainty flag,
and a one-sentence rationale. At test time we disable sampling and average each
numeric field over the 50 examples.

\FloatBarrier

\subsection{Cinematic Editing of Everyday Videos}
\label{sec:everyday-video-editing}

\begin{figure}[htbp]
    \centering
    \includegraphics[width=\linewidth]
    {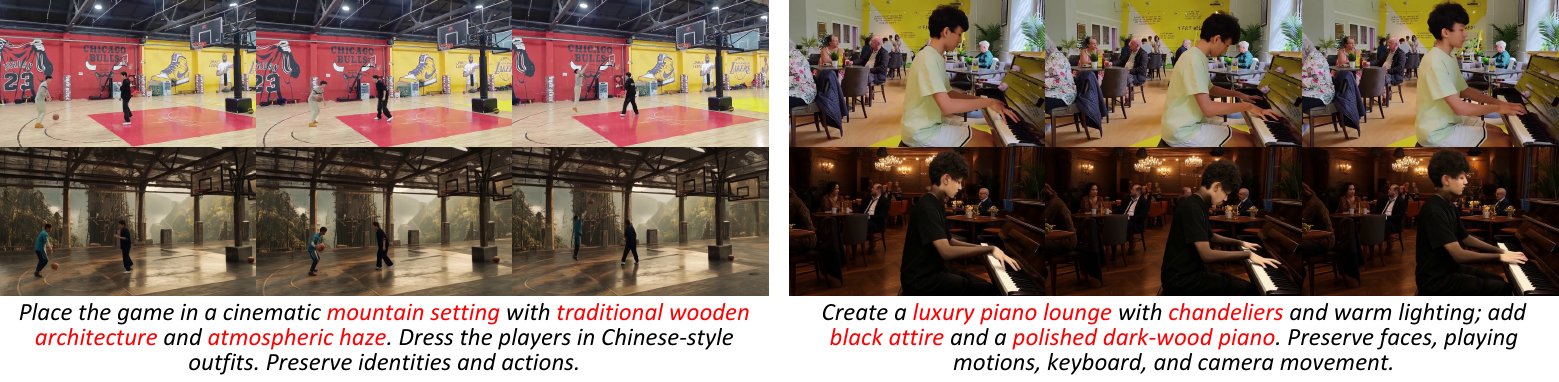}
    \caption{Cinematic editing of everyday videos.
    Basketball and piano examples show three temporally corresponding
    frames from the input videos (top) and FilmTogether outputs (bottom).
    Editing prompts are provided below each example, with key
    transformation attributes highlighted in red.}
    \label{fig:everyday-video-editing}
\end{figure}

We further illustrate FilmTogether on everyday basketball and piano videos.
Figure~\ref{fig:everyday-video-editing} shows how the model transforms
ordinary environments into cinematic settings through coordinated changes
to architecture, lighting, clothing, and materials. The basketball example
introduces a mountainous setting with wooden architecture and atmospheric
haze, while the piano example transforms a casual venue into an elegant
lounge with chandeliers and warm lighting. Three corresponding frames
illustrate these appearance changes over time.

\FloatBarrier

\Needspace{0.45\textheight}
\subsection{Memory Learning Dynamics}
\label{sec:appendix-memory-learning-dynamics}

The trajectories diagnose code differentiation and parameter adaptation rather
than generation quality. They complement the performance-based ablations in the
main paper by showing how the two memory banks evolve during continuation
training.

\begin{figure}[H]
    \centering
    \includegraphics[width=0.95\textwidth]{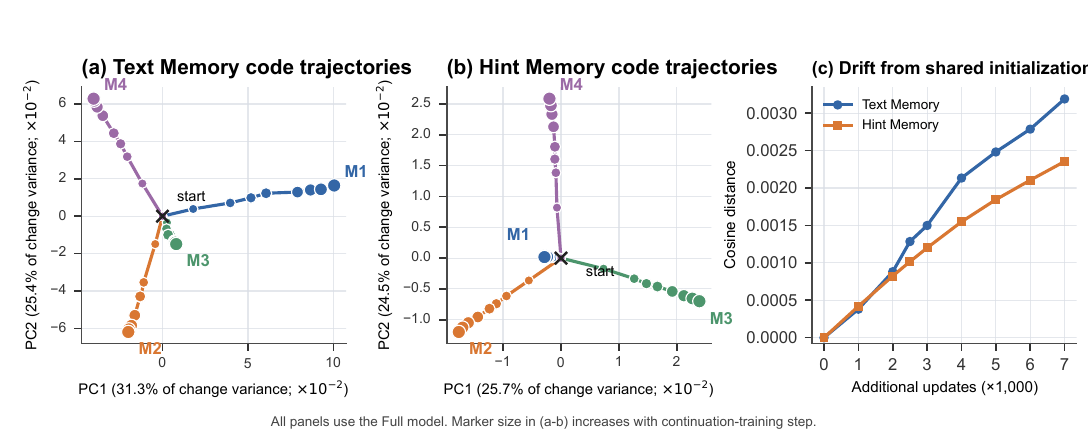}
    \vspace{-1mm}
    \caption{Learning dynamics of the dual-memory adapter. (a--b) PCA
    trajectories of the four Text and Hint Memory codes during continuation
    training of the Full model; marker size increases with optimizer updates.
    (c) Cosine distance of each complete codebook from the shared Table-1
    initialization.}
    \label{fig:memory-learning-dynamics}
\end{figure}

\FloatBarrier


\begin{thebibliography}{49}
\providecommand{\natexlab}[1]{#1}
\providecommand{\url}[1]{\texttt{#1}}
\expandafter\ifx\csname urlstyle\endcsname\relax
  \providecommand{\doi}[1]{doi: #1}\else
  \providecommand{\doi}{doi: \begingroup \urlstyle{rm}\Url}\fi

\bibitem[Ba et~al.(2016)Ba, Kiros, and Hinton]{ba2016layer}
Jimmy~Lei Ba, Jamie~Ryan Kiros, and Geoffrey~E Hinton.
\newblock Layer normalization.
\newblock \emph{arXiv preprint arXiv:1607.06450}, 2016.

\bibitem[Bai et~al.(2026)Bai, Wang, Ouyang, Yu, Wang, Wang, Cheng, Ma, Zeng, Liu, et~al.]{bai2025ditto}
Qingyan Bai, Qiuyu Wang, Hao Ouyang, Yue Yu, Hanlin Wang, Wen Wang, Ka~Leong Cheng, Shuailei Ma, Yanhong Zeng, Zichen Liu, et~al.
\newblock Scaling instruction-based video editing with a high-quality synthetic dataset.
\newblock In \emph{Proceedings of the IEEE/CVF Conference on Computer Vision and Pattern Recognition}, pp.\  37971--37981, 2026.

\bibitem[Bai et~al.(2025{\natexlab{a}})Bai, Cai, Chen, Chen, Chen, Cheng, Deng, Ding, Gao, Ge, et~al.]{bai2025qwen3vl}
Shuai Bai, Yuxuan Cai, Ruizhe Chen, Keqin Chen, Xionghui Chen, Zesen Cheng, Lianghao Deng, Wei Ding, Chang Gao, Chunjiang Ge, et~al.
\newblock Qwen3-vl technical report.
\newblock \emph{arXiv preprint arXiv:2511.21631}, 2025{\natexlab{a}}.

\bibitem[Bai et~al.(2025{\natexlab{b}})Bai, Chen, Liu, Wang, Ge, Song, Dang, Wang, Wang, Tang, et~al.]{bai2025qwen25vl}
Shuai Bai, Keqin Chen, Xuejing Liu, Jialin Wang, Wenbin Ge, Sibo Song, Kai Dang, Peng Wang, Shijie Wang, Jun Tang, et~al.
\newblock Qwen2.5-vl technical report.
\newblock \emph{arXiv preprint arXiv:2502.13923}, 2025{\natexlab{b}}.

\bibitem[Bain et~al.(2021)Bain, Nagrani, Varol, and Zisserman]{bain2021frozen}
Max Bain, Arsha Nagrani, G{\"u}l Varol, and Andrew Zisserman.
\newblock Frozen in time: A joint video and image encoder for end-to-end retrieval.
\newblock In \emph{2021 IEEE/CVF International Conference on Computer Vision (ICCV)}, pp.\  1708--1718. IEEE, 2021.

\bibitem[Caron et~al.(2021)Caron, Touvron, Misra, J{\'e}gou, Mairal, Bojanowski, and Joulin]{caron2021emerging}
Mathilde Caron, Hugo Touvron, Ishan Misra, Herv{\'e} J{\'e}gou, Julien Mairal, Piotr Bojanowski, and Armand Joulin.
\newblock Emerging properties in self-supervised vision transformers.
\newblock In \emph{2021 IEEE/CVF international conference on computer vision (ICCV)}, pp.\  9630--9640. IEEE, 2021.

\bibitem[Castellano()]{castellano_pyscenedetect}
Brandon Castellano.
\newblock {PySceneDetect}.
\newblock Software.
\newblock URL \url{https://github.com/Breakthrough/PySceneDetect}.

\bibitem[Ceylan et~al.(2023)Ceylan, Huang, and Mitra]{ceylan2023pix2video}
Duygu Ceylan, Chun-Hao~P Huang, and Niloy~J Mitra.
\newblock Pix2video: Video editing using image diffusion.
\newblock In \emph{2023 IEEE/CVF International Conference on Computer Vision (ICCV)}, pp.\  23149--23160. IEEE, 2023.

\bibitem[Chen et~al.(2025)Chen, Guo, Zhu, Zhang, Huang, Feng, and Kang]{chen2025videodepthanything}
Sili Chen, Hengkai Guo, Shengnan Zhu, Feihu Zhang, Zilong Huang, Jiashi Feng, and Bingyi Kang.
\newblock Video depth anything: Consistent depth estimation for super-long videos.
\newblock In \emph{2025 IEEE/CVF Conference on Computer Vision and Pattern Recognition (CVPR)}, pp.\  22831--22840. IEEE, 2025.

\bibitem[Chen et~al.(2024)Chen, Siarohin, Menapace, Deyneka, Chao, Jeon, Fang, Lee, Ren, Yang, et~al.]{chen2024panda}
Tsai-Shien Chen, Aliaksandr Siarohin, Willi Menapace, Ekaterina Deyneka, Hsiang-wei Chao, Byung~Eun Jeon, Yuwei Fang, Hsin-Ying Lee, Jian Ren, Ming-Hsuan Yang, et~al.
\newblock Panda-70m: Captioning 70m videos with multiple cross-modality teachers.
\newblock In \emph{2024 IEEE/CVF Conference on Computer Vision and Pattern Recognition (CVPR)}, pp.\  13320--13331. IEEE, 2024.

\bibitem[Cheng et~al.(2024)Cheng, Xiao, and He]{cheng2024consistent}
Jiaxin Cheng, Tianjun Xiao, and Tong He.
\newblock Consistent video-to-video transfer using synthetic dataset.
\newblock In \emph{International Conference on Learning Representations}, volume 2024, pp.\  16867--16879, 2024.

\bibitem[Cong et~al.(2024)Cong, Xu, Simon, Chen, Ren, Xie, Perez-Rua, Rosenhahn, Xiang, and He]{cong2024flatten}
Yuren Cong, Mengmeng Xu, Christian Simon, Shoufa Chen, Jiawei Ren, Yanping Xie, Juan-Manuel Perez-Rua, Bodo Rosenhahn, Tao Xiang, and Sen He.
\newblock Flatten: optical flow-guided attention for consistent text-to-video editing.
\newblock In \emph{International Conference on Learning Representations}, volume 2024, pp.\  52086--52106, 2024.

\bibitem[Duan et~al.(2025)Duan, Hu, Wang, Yang, Xu, Liu, Min, Cai, Ye, Zhang, et~al.]{duan2025finevq}
Huiyu Duan, Qiang Hu, Jiarui Wang, Liu Yang, Zitong Xu, Lu~Liu, Xiongkuo Min, Chunlei Cai, Tianxiao Ye, Xiaoyun Zhang, et~al.
\newblock Finevq: Fine-grained user generated content video quality assessment.
\newblock In \emph{2025 IEEE/CVF Conference on Computer Vision and Pattern Recognition (CVPR)}, pp.\  3206--3217. IEEE, 2025.

\bibitem[Geyer et~al.(2024)Geyer, Bar~Tal, Bagon, and Dekel]{geyer2024tokenflow}
Michal Geyer, Omer Bar~Tal, Shai Bagon, and Tali Dekel.
\newblock Tokenflow: Consistent diffusion features for consistent video editing.
\newblock In \emph{International Conference on Learning Representations}, volume 2024, pp.\  1608--1620, 2024.

\bibitem[He et~al.(2025)He, Wang, Zhang, Xue, Bu, Yang, Wen, and Xie]{openve2025openve}
Haoyang He, Jie Wang, Jiangning Zhang, Zhucun Xue, Xingyuan Bu, Qiangpeng Yang, Shilei Wen, and Lei Xie.
\newblock Openve-3m: A large-scale high-quality dataset for instruction-guided video editing.
\newblock \emph{arXiv preprint arXiv:2512.07826}, 2025.

\bibitem[Hendrycks \& Gimpel(2016)Hendrycks and Gimpel]{hendrycks2016gelu}
Dan Hendrycks and Kevin Gimpel.
\newblock Gaussian error linear units (gelus).
\newblock \emph{arXiv preprint arXiv:1606.08415}, 2016.

\bibitem[Hinton et~al.(2015)Hinton, Vinyals, and Dean]{hinton2015distilling}
Geoffrey Hinton, Oriol Vinyals, and Jeff Dean.
\newblock Distilling the knowledge in a neural network.
\newblock \emph{arXiv preprint arXiv:1503.02531}, 2015.

\bibitem[Ho et~al.(2020)Ho, Jain, and Abbeel]{ho2020ddpm}
Jonathan Ho, Ajay Jain, and Pieter Abbeel.
\newblock Denoising diffusion probabilistic models.
\newblock \emph{Advances in neural information processing systems}, 33:\penalty0 6840--6851, 2020.

\bibitem[Ho et~al.(2022)Ho, Salimans, Gritsenko, Chan, Norouzi, and Fleet]{ho2022videodiffusion}
Jonathan Ho, Tim Salimans, Alexey Gritsenko, William Chan, Mohammad Norouzi, and David~J Fleet.
\newblock Video diffusion models.
\newblock \emph{Advances in neural information processing systems}, 35:\penalty0 8633--8646, 2022.

\bibitem[Hu et~al.(2021)Hu, Shen, Wallis, Allen-Zhu, Li, Wang, Wang, and Chen]{hu2022lora}
Edward~J Hu, Yelong Shen, Phillip Wallis, Zeyuan Allen-Zhu, Yuanzhi Li, Shean Wang, Lu~Wang, and Weizhu Chen.
\newblock Lora: Low-rank adaptation of large language models.
\newblock \emph{arXiv preprint arXiv:2106.09685}, 2021.

\bibitem[Huang et~al.(2026)Huang, Xu, Luo, Hu, Tang, Peng, Zhang, Wang, and Fu]{huang2026ffp300k}
Xijie Huang, Chengming Xu, Donghao Luo, Xiaobin Hu, Peng Tang, Xu~Peng, Jiangning Zhang, Chengjie Wang, and Yanwei Fu.
\newblock Ffp-300k: Scaling first-frame propagation for generalizable video editing.
\newblock In \emph{Proceedings of the IEEE/CVF Conference on Computer Vision and Pattern Recognition}, pp.\  23172--23181, 2026.

\bibitem[Huang et~al.(2024)Huang, He, Yu, Zhang, Si, Jiang, Zhang, Wu, Jin, Chanpaisit, et~al.]{huang2024vbench}
Ziqi Huang, Yinan He, Jiashuo Yu, Fan Zhang, Chenyang Si, Yuming Jiang, Yuanhan Zhang, Tianxing Wu, Qingyang Jin, Nattapol Chanpaisit, et~al.
\newblock Vbench: Comprehensive benchmark suite for video generative models.
\newblock In \emph{2024 IEEE/CVF Conference on Computer Vision and Pattern Recognition (CVPR)}, pp.\  21807--21818. IEEE, 2024.

\bibitem[Jiang et~al.(2025)Jiang, Han, Mao, Zhang, Pan, and Liu]{jiang2025vace}
Zeyinzi Jiang, Zhen Han, Chaojie Mao, Jingfeng Zhang, Yulin Pan, and Yu~Liu.
\newblock Vace: All-in-one video creation and editing.
\newblock In \emph{2025 IEEE/CVF International Conference on Computer Vision (ICCV)}, pp.\  17191--17202. IEEE, 2025.

\bibitem[Khachatryan et~al.(2023)Khachatryan, Movsisyan, Tadevosyan, Henschel, Wang, Navasardyan, and Shi]{khachatryan2023text2videozero}
Levon Khachatryan, Andranik Movsisyan, Vahram Tadevosyan, Roberto Henschel, Zhangyang Wang, Shant Navasardyan, and Humphrey Shi.
\newblock Text2video-zero: Text-to-image diffusion models are zero-shot video generators.
\newblock In \emph{2023 IEEE/CVF International Conference on Computer Vision (ICCV)}, pp.\  15908--15918. IEEE, 2023.

\bibitem[Kirstain et~al.(2023)Kirstain, Polyak, Singer, Matiana, Penna, and Levy]{kirstain2023pick}
Yuval Kirstain, Adam Polyak, Uriel Singer, Shahbuland Matiana, Joe Penna, and Omer Levy.
\newblock Pick-a-pic: An open dataset of user preferences for text-to-image generation.
\newblock \emph{Advances in neural information processing systems}, 36:\penalty0 36652--36663, 2023.

\bibitem[Ku et~al.(2024)Ku, Wei, Ren, Yang, and Chen]{ku2024anyv2v}
Max Ku, Cong Wei, Weiming Ren, Harry Yang, and Wenhu Chen.
\newblock Anyv2v: A tuning-free framework for any video-to-video editing tasks.
\newblock \emph{arXiv preprint arXiv:2403.14468}, 2024.

\bibitem[Liao et~al.(2025)Liao, Zeng, Song, Fu, Yu, and Lin]{liao2025context}
Xinyao Liao, Xianfang Zeng, Ziye Song, Zhoujie Fu, Gang Yu, and Guosheng Lin.
\newblock In-context learning with unpaired clips for instruction-based video editing.
\newblock \emph{arXiv preprint arXiv:2510.14648}, 2025.

\bibitem[Lin et~al.(2026)Lin, Liang, Zeng, Bai, Chen, and Shou]{kiwiedit}
Yiqi Lin, Guoqiang Liang, Ziyun Zeng, Zechen Bai, Yanzhe Chen, and Mike~Zheng Shou.
\newblock Kiwi-edit: Versatile video editing via instruction and reference guidance.
\newblock \emph{arXiv preprint arXiv:2603.02175}, 2026.

\bibitem[Liu et~al.(2024)Liu, Zhang, Li, Lin, and Jia]{liu2024videop2p}
Shaoteng Liu, Yuechen Zhang, Wenbo Li, Zhe Lin, and Jiaya Jia.
\newblock Video-p2p: Video editing with cross-attention control.
\newblock In \emph{2024 IEEE/CVF Conference on Computer Vision and Pattern Recognition (CVPR)}, pp.\  8599--8608. IEEE, 2024.

\bibitem[Loshchilov \& Hutter(2017)Loshchilov and Hutter]{loshchilov2019decoupled}
Ilya Loshchilov and Frank Hutter.
\newblock Decoupled weight decay regularization.
\newblock \emph{arXiv preprint arXiv:1711.05101}, 2017.

\bibitem[Nagrani et~al.(2022)Nagrani, Seo, Seybold, Hauth, Manen, Sun, and Schmid]{nagrani2022learning}
Arsha Nagrani, Paul~Hongsuck Seo, Bryan Seybold, Anja Hauth, Santiago Manen, Chen Sun, and Cordelia Schmid.
\newblock Learning audio-video modalities from image captions.
\newblock In \emph{European Conference on Computer Vision}, pp.\  407--426. Springer, 2022.

\bibitem[Nan et~al.(2025)Nan, Xie, Zhou, Fan, Yang, Chen, Li, Yang, and Tai]{nan2025openvid}
Kepan Nan, Rui Xie, Penghao Zhou, Tiehan Fan, Zhenheng Yang, Zhijie Chen, Xiang Li, Jian Yang, and Ying Tai.
\newblock Openvid-1m: A large-scale high-quality dataset for text-to-video generation.
\newblock In \emph{International conference on learning representations}, volume 2025, pp.\  1045--1064, 2025.

\bibitem[Peebles \& Xie(2023)Peebles and Xie]{peebles2023dit}
William Peebles and Saining Xie.
\newblock Scalable diffusion models with transformers.
\newblock In \emph{2023 IEEE/CVF International Conference on Computer Vision (ICCV)}, pp.\  4172--4182. IEEE, 2023.

\bibitem[Qi et~al.(2023)Qi, Cun, Zhang, Lei, Wang, Shan, and Chen]{qi2023fatezero}
Chenyang Qi, Xiaodong Cun, Yong Zhang, Chenyang Lei, Xintao Wang, Ying Shan, and Qifeng Chen.
\newblock Fatezero: Fusing attentions for zero-shot text-based video editing.
\newblock In \emph{2023 IEEE/CVF International Conference on Computer Vision (ICCV)}, pp.\  15886--15896. IEEE, 2023.

\bibitem[Radford et~al.(2021)Radford, Kim, Hallacy, Ramesh, Goh, Agarwal, Sastry, Askell, Mishkin, Clark, et~al.]{radford2021learning}
Alec Radford, Jong~Wook Kim, Chris Hallacy, Aditya Ramesh, Gabriel Goh, Sandhini Agarwal, Girish Sastry, Amanda Askell, Pamela Mishkin, Jack Clark, et~al.
\newblock Learning transferable visual models from natural language supervision.
\newblock In \emph{International conference on machine learning}, pp.\  8748--8763. PmLR, 2021.

\bibitem[Rombach et~al.(2022)Rombach, Blattmann, Lorenz, Esser, and Ommer]{rombach2022latent}
Robin Rombach, Andreas Blattmann, Dominik Lorenz, Patrick Esser, and Bj{\"o}rn Ommer.
\newblock High-resolution image synthesis with latent diffusion models.
\newblock In \emph{2022 IEEE/CVF conference on computer vision and pattern recognition (CVPR)}, pp.\  10674--10685. ieee, 2022.

\bibitem[Song et~al.(2020{\natexlab{a}})Song, Meng, and Ermon]{song2020ddim}
Jiaming Song, Chenlin Meng, and Stefano Ermon.
\newblock Denoising diffusion implicit models.
\newblock \emph{arXiv preprint arXiv:2010.02502}, 2020{\natexlab{a}}.

\bibitem[Song et~al.(2020{\natexlab{b}})Song, Sohl-Dickstein, Kingma, Kumar, Ermon, and Poole]{song2021score}
Yang Song, Jascha Sohl-Dickstein, Diederik~P Kingma, Abhishek Kumar, Stefano Ermon, and Ben Poole.
\newblock Score-based generative modeling through stochastic differential equations.
\newblock \emph{arXiv preprint arXiv:2011.13456}, 2020{\natexlab{b}}.

\bibitem[Team et~al.(2026)Team, Liu, Chen, Li, Chi, Sun, Li, Fu, Guo, Wu, et~al.]{team2026bernini}
Bernini Team, Chenchen Liu, Junyi Chen, Lei Li, Lu~Chi, Mingzhen Sun, Zhuoying Li, Yi~Fu, Ruoyu Guo, Yiheng Wu, et~al.
\newblock Bernini: Latent semantic planning for video diffusion.
\newblock \emph{arXiv preprint arXiv:2605.22344}, 2026.

\bibitem[Team(2025)]{decart2025lucyedit}
DecartAI Team.
\newblock Lucy edit: Open-weight text-guided video editing, 2025.

\bibitem[Vaswani et~al.(2017)Vaswani, Shazeer, Parmar, Uszkoreit, Jones, Gomez, Kaiser, and Polosukhin]{vaswani2017attention}
Ashish Vaswani, Noam Shazeer, Niki Parmar, Jakob Uszkoreit, Llion Jones, Aidan~N Gomez, {\L}ukasz Kaiser, and Illia Polosukhin.
\newblock Attention is all you need.
\newblock \emph{Advances in neural information processing systems}, 30, 2017.

\bibitem[Wan et~al.(2025)Wan, Wang, Ai, Wen, Mao, Xie, Chen, Yu, Zhao, Yang, et~al.]{wan2025}
Team Wan, Ang Wang, Baole Ai, Bin Wen, Chaojie Mao, Chen-Wei Xie, Di~Chen, Feiwu Yu, Haiming Zhao, Jianxiao Yang, et~al.
\newblock Wan: Open and advanced large-scale video generative models.
\newblock \emph{arXiv preprint arXiv:2503.20314}, 2025.

\bibitem[Wang et~al.(2024)Wang, He, Li, Li, Yu, Ma, Li, Chen, Chen, Wang, et~al.]{wang2024internvid}
Yi~Wang, Yinan He, Yizhuo Li, Kunchang Li, Jiashuo Yu, Xin Ma, Xinhao Li, Guo Chen, Xinyuan Chen, Yaohui Wang, et~al.
\newblock Internvid: A large-scale video-text dataset for multimodal understanding and generation.
\newblock In \emph{International Conference on Learning Representations}, volume 2024, pp.\  42055--42079, 2024.

\bibitem[Wu et~al.(2025{\natexlab{a}})Wu, Li, Zhou, Lin, Gao, Yan, Yin, Bai, Xu, Chen, et~al.]{wu2025qwenimage}
Chenfei Wu, Jiahao Li, Jingren Zhou, Junyang Lin, Kaiyuan Gao, Kun Yan, Sheng-ming Yin, Shuai Bai, Xiao Xu, Yilei Chen, et~al.
\newblock Qwen-image technical report.
\newblock \emph{arXiv preprint arXiv:2508.02324}, 2025{\natexlab{a}}.

\bibitem[Wu et~al.(2023)Wu, Ge, Wang, Lei, Gu, Shi, Hsu, Shan, Qie, and Shou]{wu2023tuneavideo}
Jay~Zhangjie Wu, Yixiao Ge, Xintao Wang, Stan~Weixian Lei, Yuchao Gu, Yufei Shi, Wynne Hsu, Ying Shan, Xiaohu Qie, and Mike~Zheng Shou.
\newblock Tune-a-video: One-shot tuning of image diffusion models for text-to-video generation.
\newblock In \emph{2023 IEEE/CVF International Conference on Computer Vision (ICCV)}, pp.\  7589--7599. IEEE, 2023.

\bibitem[Wu et~al.(2025{\natexlab{b}})Wu, Chen, Li, Wang, Xie, and Zhang]{wu2025insvie}
Yuhui Wu, Liyi Chen, Ruibin Li, Shihao Wang, Chenxi Xie, and Lei Zhang.
\newblock Insvie-1m: Effective instruction-based video editing with elaborate dataset construction.
\newblock In \emph{2025 IEEE/CVF International Conference on Computer Vision (ICCV)}, pp.\  16692--16701. IEEE, 2025{\natexlab{b}}.

\bibitem[Zhang et~al.(2025)Zhang, Long, Li, Qiu, Liu, Yao, and Mei]{zhang2025reco}
Zhongwei Zhang, Fuchen Long, Wei Li, Zhaofan Qiu, Wu~Liu, Ting Yao, and Tao Mei.
\newblock Region-constraint in-context generation for instructional video editing.
\newblock \emph{arXiv preprint arXiv:2512.17650}, 2025.

\bibitem[Zhao et~al.(2026)Zhao, Pan, He, Yu, Chen, Ye, Liu, Xie, and Han]{zhao2026sana}
Yuyang Zhao, Yicheng Pan, Qiyuan He, Jincheng Yu, Junsong Chen, Tian Ye, Haozhe Liu, Enze Xie, and Song Han.
\newblock Sana-streaming: Real-time streaming video editing with hybrid diffusion transformer.
\newblock \emph{arXiv preprint arXiv:2605.30409}, 2026.

\bibitem[Zi et~al.(2026)Zi, Ruan, Chen, Qi, Hao, Zhao, Huang, Liang, Xiao, and Wong]{zi2025senorita}
Bojia Zi, Penghui Ruan, Marco Chen, Xianbiao Qi, Shaozhe Hao, Shihao Zhao, Youze Huang, Bin Liang, Rong Xiao, and Kam-Fai Wong.
\newblock Se{\~n}orita-2m: A high-quality instruction-based dataset for general video editing by video specialists.
\newblock \emph{Advances in Neural Information Processing Systems}, 38, 2026.

\end{thebibliography}
\end{document}